\documentclass{llncs}
\usepackage{bm}
\usepackage[utf8]{inputenc}
\usepackage{url}
\usepackage{amssymb}
\usepackage{amsfonts}
\usepackage{graphicx}
\usepackage{url}
\usepackage{lstsemantic}
\usepackage{lstlangmizar}
\usepackage{booktabs}
\usepackage{footnoterange}
\usepackage{cite}

\usepackage{float}
\floatstyle{boxed}

\usepackage{hyperref}

\usepackage{xcolor}

\title{ENIGMAWatch: ProofWatch Meets ENIGMA}

\author{Zarathustra Goertzel \and Jan Jakub\r{u}v
  \and Josef Urban\thanks{Supported by the \textit{AI4REASON} ERC Consolidator grant number 649043, and by the Czech project AI\&Reasoning CZ.02.1.01/0.0/0.0/15\_003/0000466 and the European Regional Development Fund.}}
\institute{
  Czech Technical University in Prague
}

\authorrunning{Goertzel, Jakub\r{u}v, Urban}

\titlerunning{ProofWatch Meets ENIGMA}

\newcommand*{\progress}{\mathop{\mathit{progress}}}

\newcommand*{\stdrel}{\mathop{\mathit{relevance}}}

\newcommand*{\M}{\mathcal{M}}
\newcommand{\Clauses}{\mathcal{C}}
\newcommand{\PosC}{\Clauses^{+}}
\newcommand{\NegC}{\Clauses^{-}}

\makeatletter
\renewcommand\section{\@startsection{section}{1}{\z@}
                       {-12\p@ \@plus -4\p@ \@minus -4\p@}
                       {8\p@ \@plus 4\p@ \@minus 4\p@}
                       {\normalfont\large\bfseries\boldmath
                        \rightskip=\z@ \@plus 8em\pretolerance=10000 }}
\makeatother

\begin{document}

\maketitle{}

\begin{abstract}
  In this work we describe a new learning-based proof guidance --
  ENIGMAWatch -- for saturation-style first-order theorem provers.
  ENIGMAWatch combines two guiding approaches for the given-clause
  selection implemented for the E ATP system: ProofWatch and ENIGMA. 
  ProofWatch is motivated by the watchlist (hints) method and based on symbolic
  matching of multiple related proofs, while ENIGMA is based on
  statistical machine learning. The two methods are combined by using
  the evolving information about symbolic proof matching as 
  additional characterization of the saturation-style proof
  search for the statistical learning methods.  The new system is
  evaluated on a large set of problems from the Mizar
  library. We show that the added proof-matching information is
  considered important by the statistical machine learners, and that
  it leads to improved performance over ProofWatch and
  ENIGMA.

\end{abstract}

\section{Introduction}
This work describes a new learning-based proof guidance --
\emph{ENIGMAWatch} -- for saturation-style first-order theorem
provers. ENIGMAWatch
\footnote{The E version used in this paper can be found at \url{https://github.com/ai4reason/eprover/tree/devel},
and the library for running ENIGMA with E can be found at \url{https://github.com/ai4reason/enigma}.}
 is the combination of two previous guidance methods
implemented for the E theorem prover~\cite{Schulz13}:
ProofWatch~\cite{DBLP:conf/itp/GoertzelJ0U18} and
ENIGMA~\cite{JakubuvU17a,JakubuvU18}. 
Both ProofWatch and ENIGMA learn to guide E's proof search for a new
conjecture based on related proofs.

ProofWatch uses the hints (watchlist) mechanism, which is a form of 
precise symbolic 
memory that can allow inference chains done in a former
proof to be replayed in the current proof search. It uses standard symbolic 
subsumption to check which clauses subsume 
clauses in related proofs.
In addition to boosting the priority of these clauses, the completion ratios
of the related proofs are computed, and the proof search is biased towards the 
most completed ones.

ENIGMA uses fast statistical machine learning to learn from related
proof-searches to identify good and bad (positive and negative)
clauses for the current conjecture. ENIGMA chooses the given clauses
based only on features of the problem's conjecture, which is static
throughout the whole proof search. This seems suboptimal: as the proof
search evolves, information about the work done so far should
influence the selection of the next given clauses.

ENIGMAWatch combines the two approaches by giving the ENIGMA's learner the ProofWatch completion ratios of 
the related proofs as an evolving vectorial characterization of the current proof search state. 
This allows E's machine learning guidance to
have more information about how the proof search is unfolding.

An early version of ENIGMAWatch was tested on the MPTP Challenge\footnote{\url{http://tptp.cs.miami.edu/~tptp/MPTPChallenge/}}~\cite{Urban06,UrbanMPTPChallenge} 
benchmark. It contains 252 first-order problems
extracted from the Mizar Mathematical Library (MML)~\cite{mizar-in-a-nutshell}, 
used in Mizar to prove the Bolzano-Weierstrass theorem.
Initially, ENIGMAWatch could not be run on a larger dataset, such as the $57897$ 
Mizar40~\cite{KaliszykU13b} benchmark, in a reasonable time.
Since then, ENIGMA implemented dimensionality reduction using feature
hashing~\cite{DBLP:journals/corr/abs-1903-03182}, extending its applicability to
large corpora.
We have additionally improved watchlist mechanism in E through enhanced
indexing, first time presented in this work in Section~\ref{Indexing}.
This allows also ENIGMAWatch to be applied to larger corpora.

The rest of the paper is organized as follows. 
Section~\ref{ATP} provides an introduction to saturation-based theorem proving
and briefly describes ENIGMA and ProofWatch.
Section~\ref{ENIGMAWatch} explains how ENIGMA and
ProofWatch are combined into ENIGMAWatch, and how watchlists can be selected. 
Section~\ref{Indexing} describes our improved watchlist indexing in E.
Both ENIGMAWatch and the improved watchlist indexing are evaluated in
Section~\ref{Eval}.

\section{Guiding the Given Clause Selection in ATPs}
\label{ATP}

\subsection{Automated Theorem Proving and Machine Learning} 
\label{Sat}
State-of-the-art saturation-based automated theorem provers (ATPs) for
first-order logic (FOL), such as E~\cite{Sch02-AICOMM} and Vampire~\cite{Vampire}
employ the \emph{given clause algorithm}, translating
the input FOL problem $T\cup\{\lnot C\}$ into a refutationally
equivalent set of clauses.
The search for a contradiction is performed maintaining sets of
\emph{processed} ($P$) and \emph{unprocessed} ($U$) clauses (the \emph{proof state} $\Pi$).
The algorithm repeatedly selects a \emph{given clause} $g$ from $U$,
moves $g$ to $P$, and extends $U$ with all clauses inferred with $g$ and $P$.
This process continues until a contradiction is found, $U$ becomes empty, or 
a resource limit is reached.

The search space of this loop grows quickly and it is a well-known fact that the
selection of the right given clause is crucial for success.
Machine learning from a large number of proofs and proof searches~~\cite{DBLP:conf/ogai/ErtelSS89,DenzingerFGS99,DBLP:books/daglib/0002958,US+08-long,abs-1108-3446,UrbanVS11,holyhammer,DBLP:journals/jar/BridgeHP14,blistr,KaliszykU15,SchaferS15,DBLP:conf/cade/FarberB16,BlanchetteGKKU16,hh4h4,IrvingSAECU16,JakubuvU17a,LoosISK17,PiotrowskiU18,JakubuvU18a,KaliszykUMO18} may help guide
the selection of the given clauses.

\subsection{ENIGMA: Learning from Successful Proof Searches}
\label{Enigma}

ENIGMA~\cite{JakubuvU17a,JakubuvU18,DBLP:journals/corr/abs-1903-03182,DBLP:journals/corr/abs-1904-01677}  
(\emph{Efficient learNing-based Internal Guidance MAchine}) is our method for 
guiding given clause selection in saturation-based ATPs.
The method needs to be efficient because it is internally applied to every generated clause. 
ENIGMA uses E's capability to analyze
successful proof searches, and to output lists of given clauses annotated as either
\emph{positive} or \emph{negative} training examples.
Each processed clause which is present in the final proof is classified as
positive. 
On the other hand, processing of clauses not present in the final proof was redundant,
hence they are classified as negative.
ENIGMA's goal is to learn such classification (possibly conditioned on the problem
and its features) in a way that generalizes and allows solving new related
problems.

\newcommand*{\T}{\mathcal{T}}
\renewcommand*{\P}{\mathcal{P}}
\renewcommand*{\S}{\mathcal{S}}

\noindent\textbf{ENIGMA Learning and Models.}
Given a set of problems $\P$, we can run E with a strategy $\S$ and obtain positive and
negative training data
$\T$ from each of the successful proof searches.
Various machine learning methods can be used to learn the clause classification
given by $\T$, each method yielding a \emph{classifier} or a (classification) \emph{model} $\M$.
In order to use the model $\M$ in E, $\M$ is used as a function that
computes clause weights.
This weight function is then used to guide future E runs.

First-order clauses need to be represented in a format recognized by the
selected learning method.
While neural networks have been very recently practically used for
internal guidance with ENIGMA~\cite{DBLP:journals/corr/abs-1903-03182}, the strongest setting currently uses
manually engineered \emph{clause features} and fast non-neural state-of-the-art
gradient boosted trees libraries such as XGBoost~\cite{DBLP:conf/kdd/ChenG16}.
The model $\M$ produced by XGBoost consists of a set (\emph{ensemble}~\cite{Polikar06}) of decision trees.
Given a clause $C$, the model $\M$ yields the probability that $C$ represents a
positive clause.
When using $\M$ as a weight function in E, the probabilities are turned into binary
classification, assigning weight $1.0$ for probabilities $\ge 0.5$ and
weight $10.0$ otherwise.

\noindent\textbf{Clause Features.}
Clause features represent a finite set of various syntactic properties of
clauses, and are used to encode clauses by a fixed-length numeric vector.
Various machine learning methods can handle numeric vectors and their success
heavily depends on the selection of correct clause features.
Various possible choices of efficient clause features for theorem prover
guidance have been experimented
with~\cite{JakubuvU17a,JakubuvU18,KaliszykUMO18,DBLP:conf/ijcai/KaliszykUV15}.
The original ENIGMA~\cite{JakubuvU17a} uses term-tree walks of
length 3 as features, while the second version~\cite{JakubuvU18} reaches better
results by employing various additional features.

Since there are only finitely many features in any training data, the features
can be serially numbered.
This numbering is fixed for each experiment.
Let $n$ be the number of different features appearing in the training data.
A clause $C$ is translated to a feature vector
$\varphi_C$ whose $i$-th member counts the number of occurrences of the $i$-th 
feature in $C$.
Hence every clause is represented by a sparse numeric vector of length $n$.
Additionally, we embed information about the conjecture currently being proved
in the feature vector, yielding vectors of length $2n$.
See \cite{DBLP:journals/corr/abs-1903-03182,JakubuvU18} for more details.

\noindent\textbf{Feature Hashing.}
Experiments revealed that XGBoost is capable of
dealing with vectors up to the length of $10^5$ with a reasonable performance.
In experiments with
the whole translated Mizar Mathematical Library, the feature vector
length can easily grow over $10^6$.  This significantly increases both
the training and the clause evaluation times.  To handle such larger
data sets, a simple \emph{hashing} method  has previously been implemented to
decrease the dimension of the vectors.

Instead of serially numbering all features, 
we represent
each feature $f$ by a unique string and apply a general-purpose string
hashing function 
to obtain a number $n_f$ within a required range (between 0 and an
adjustable \emph{hash base}).
The value of $f$ is then stored in the feature vector at the position $n_f$.
If different features get mapped to the same
vector index, the corresponding values are summed up.
See \cite{DBLP:journals/corr/abs-1903-03182} for more details.

\subsection{ProofWatch: Proof Guidance by Clause Subsumption}
\label{ProofWatch}

In this section we explain the ProofWatch guiding mechanisms. 
Unlike the statistical approach in ENIGMA, ProofWatch implements a form of \emph{symbolic} memory and guidance. 
It produces a notion of \emph{proof-state vector} that is dynamically created and updated.

\noindent\textbf{Standard Watchlist Guidance.}
The watchlist (hint list) mechanism itself does not perform any statistical machine learning.
It steers given clause selection via symbolic matching between
generated clauses and a set of clauses called a \emph{watchlist}. 
This technique has been originally developed by Veroff~\cite{Veroff:JAR-1996}
and implemented in Otter~\cite{MW:JAR-97} and Prover9~\cite{McCune:WWW-2008}.
Since then, it has been extensively used in the AIM project~\cite{KinyonVV13} 
for obtaining long and advanced proofs of open algebraic conjectures.  
The watchlist mechanism is nowadays implemented also in E. 
All the above implementations use only a single watchlist, as opposed to
ProofWatch discussed below.

Recall that a clause $C$ \emph{subsumes} a clause $D$, written $C \sqsubseteq
D$, when there exists a
substitution $\sigma$ such that $C\sigma\subseteq D$ (where clauses are
considered to be sets of literals).
The watchlist guidance then works as follows.
Every generated clause $C$ is checked for subsumption with every watchlist
clause $D\in W$.
When $C$ subsumes at least one of the watchlist clauses, then $C$ is considered
important for the proof search and is processed with high priority.
The idea behind this is that the watchlist $W$ contains clauses which were processed
during a previous successful proof search of a related conjecture.
Hence processing of similar clauses may lead to success again.

In E, the watchlist mechanism is implemented using a priority function
\footnote{See the priority function \texttt{PreferWatchlist} in the E manual.}
which takes precedence over the weight function used to select the next given clause.
Priority functions assign the priority to each clause, and clauses with higher
priority are selected as given before clauses with lower
priority\footnote{Numerically the lower the priority, the better. Hence $0$ is the
best priority.}.
When clauses from previous proofs are put on a watchlist, E thus prefers to
follow steps from the previous proofs whenever it can.

\noindent\textbf{ProofWatch.}
Our approach~\cite[Sec. 5]{DBLP:conf/itp/GoertzelJ0U18} extends standard watchlist 
guidance by allowing for multiple watchlists $W_1$,$\ldots$,$W_n$, for example, 
one corresponding to each related proof found before. 
We say that a generated clause $C$ \emph{matches} the watchlist $W_i$, written
$C\sqsubseteq W_i$, iff $C$ subsumes some clause $D\in W_i$ ($C\sqsubseteq D$).
Similarly, the above watchlist clause $D$ is said to be \emph{matched} by $C$.

The reason to include multiple watchlists is that during a proof search, 
clauses from some watchlists might get matched more often
than clauses from others.
The more clauses are matched from some watchlist $W_i$, 
the more the current proof search resembles $W_i$,
and hence $W_i$ might be more relevant for this proof search.
Thus the idea of ProofWatch is to prioritize clauses that match more relevant
watchlists (proofs).

Watchlist \emph{relevance} is dynamically computed as follows.
We define $\progress(W_i)$ to be the count of clauses from $W_i$ that have been
matched in the proof search thus far. 
The \emph{completion ratio}, $c_i = \frac{\progress(W_i)}{|W_i|}$, measures how much of the
watchlist $W_i$ has been matched. The \emph{dynamic relevance} of each generated
clause $C$ is defined as the maximum completion ratio over all the watchlists
$W_i$ that $C$ matches:
\[
   \stdrel(C) =
   \max_{W\in
      \{W_i: C\sqsubseteq W_i\}
   }
   \Big(
      \frac{\progress(W)}{|W|}
   \Big)
\]
The higher the dynamic relevance $\stdrel(C)$, the higher the priority of $C$.
The dynamic watchlist mechanism is implemented using the E priority
function.\footnote{See \texttt{PreferWatchlistRelevant} in \cite{DBLP:conf/itp/GoertzelJ0U18}.}
The results of experiments in \cite[Sec.~6.3]{DBLP:conf/itp/GoertzelJ0U18} on the same dataset as this work (Mizar40~\cite{KaliszykU13b}) indicate that dynamic relevance 
improves performance over an ensemble of strategies, whereas the single watchlist approach is stronger on
each individual strategy. 

When using a large problem library such as Mizar40, it is practically useful to choose only some proofs for watchlists. 
First, E's speed decreases with each additional proof on the watchlist, so if working on a large dataset,
loading all available proofs as watchlists will lead to a large slowdown (cf. Section~\ref{Indexing}).
Second, it's not guaranteed that all proofs will help E with proving the problem at hand. 

\section{ENIGMAWatch: ProofWatch meets ENIGMA}
\label{ENIGMAWatch}

\subsection{Completion Ratios as Semantic Embeddings of the Proof Search}
The watchlist completion ratios $(c_0,...,c_N)$ ($N$ ranges over the watchlist proofs) at each step in E's proof search can
be taken as a vectorial representation of the current proof state $\Pi$. The general motivation 
for this approach is to come up with an \emph{evolving} characterization of the
saturation-style proof state $\Pi$, preferably in a vectorial form $\varphi_\Pi$ suitable
for machine learning tools, such as ENIGMA. 

Recall that the proof state $\Pi$ is a set of processed clauses $P$
and unprocessed clauses  $U$. The vector of watchlist completion ratios thus maintains a running tally of 
where clauses in $P \cup U$ match the different related proofs.
In general, this could be replaced, e.g., by a vector
of more abstract similarities of the current proof state to other
proofs measured in various (possibly approximate) ways. 
In ENIGMAWatch we use the ProofWatch based \emph{proof-state vector} for a proof state $\Pi$ defined by the completion ratios, i.e., $\varphi_\Pi =(c_0,\ldots,c_N)$. 
This is the first practical implementation of the general idea: using \emph{semantic embeddings} (i.e., representations in $R^n$)  of the proof state $\Pi$ for guiding statistical learning methods.
ENIGMAWatch  uses the proof-state vectors $\varphi_\Pi$ as follows. 
The positive $\PosC$ and negative $\NegC$ given clauses are output along with $\varphi_\Pi$, 
the proof-state vector at the time of their selection, and used as added features of the proof state when training ENIGMA-style classifiers. 

Table~\ref{tabprog} shows a sample proof-state vector based on $32$ related proofs\footnote{The proofs were chosen via k-NN. 
See \cite[Sec. 6.1]{DBLP:conf/itp/GoertzelJ0U18} for details.} for the Mizar theorem 
\textbf{YELLOW 5:36}\footnote{\url{http://grid01.ciirc.cvut.cz/~mptp/7.13.01\_4.181.1147/html/yellow\_5\#T36}} 
(De Morgan's law\footnote{$\neg (P \vee Q) \iff (\neg P) \wedge (\neg Q)$})
at the end of the proof search.
Note that some related proofs, such as $\#2$, were almost fully matched, while others, such as $\#7$ were mostly not matched in the proof search.

\begin{table}[t]
\begin{footnotesize}
  \setlength\tabcolsep{5.5pt}
  \center
\begin{tabular}{ccc|ccc|ccc|ccc}
 0& 0.438 &      42/96& 1& 0.727 &      56/77& 2& 0.865 &      45/52 & 3 & 0.360 &       9/25\\
 4& 0.750 &      51/68& 5& 0.259 &       7/27& 6& 0.805 &      62/77 & 7 & 0.302 &      73/242\\
\end{tabular}
\end{footnotesize}
\caption{\label{tabprog}{
    Example of the proof-state vector for 8 (of 32) (serially numbered) proofs loaded to guide the proof of \texttt{YELLOW\_5:36}. 
The three columns are the watchlist $i$, the completion ratio of $i$, and $\progress(W_i)/|W_i|$.}}
\vspace{-6mm}
\end{table}

\subsection{Proof Vector Construction}

\noindent\textbf{Data Construction.}
In the ProofWatch~\cite{DBLP:conf/itp/GoertzelJ0U18} experiments, 
the best method for selecting related proofs (watchlists) was to use k-nearest neighbor (k-NN) 
to recommend 32 proofs per problem. The watchlists there are thus problem specific.
In ENIGMAWatch, we want the watchlists to be globally fixed across the whole library, so that the proof completion ratios
have the same meaning in all proofs.
To construct the proof vectors, we first use a strong E strategy to produce a set of initial proofs ($14882$ over the $57897$ Mizar40 problems). 
Then we run E with ProofWatch and the same strategy over the full $57897$ problems with the $14882$ proofs
loaded into the watchlist. The time limit for both runs was \emph{T60-G10000}, which means that E stops after 60 seconds
or $10000$ generated clauses.
This data provides information on how often each watchlist was encountered in each successful 
proof search.
The training data then consists of a proof vector for each given clause (for each conjecture/problem):
$(conjecture, given\mbox{-}clause, proof\mbox{-}state\ vector)$.

\noindent\textbf{Dimensionality Reduction.}
Next, we experiment with various pre-processing methods to reduce the $proof\mbox{-}state\ vector$ dimension
and thus decrease the number of watchlists loaded in E. 
For each problem we compute the mean of proof-state vectors over all given clauses $g$: 
$\frac{1}{\# g} \sum_{g} \varphi_{\Pi_g}$. 
This vector consists of the averaged completion ratios for each watchlist, 
which will be higher if the watchlist was matched earlier in the proof.
This results in the mean proof-state matrix $M$ consisting of row vectors $(mean\mbox{-}proof\mbox{-}vector)$ (one for each conjecture/problem).

The following are methods experimented with in this paper for constructing the globally fixed vector of 
$512$ watchlists from matrix $M$:
\vspace{-2mm}
\begin{itemize}
    \item \emph{Mean}: compute the mean of $M$ across the rows to obtain 
        a mean proof-state vector that contains for each watchlist its average use across all problems.
Then we take the top 
        $512$ watchlists.
    \item \emph{Corr}: compute the Pearson correlation matrix\footnote{\url{https://docs.scipy.org/doc/numpy/reference/generated/numpy.corrcoef.html}} based on (the transpose of) $M$, 
        and find a relatively uncorrelated set of $512$ watchlists.
    \item \emph{Var}: compute the variance (across the rows) of each column in $M$, and take the $512$ watchlists with the highest variance. 
        The intuition is that watchlists whose completion ratio vary more over the problem corpus may be more useful for learning.
    \item \emph{Rand}: randomly select $512$ watchlists.
\end{itemize}

\section{Multi-indices Subsumption Indexing}
\label{Indexing}

In order to determine whether a generated clause matches a watchlist, the
generated clause must be checked for subsumption with every watchlist clause.
A major limitation of previous
work~\cite{DBLP:conf/itp/GoertzelJ0U18,LPAR-IWIL2018:ProofWatch_Meets_ENIGMA_First}
was the slowdown of E as the watchlist size increased beyond
$4000$ clauses. Including more than $128$ proofs was impractical.
This section describes a method we have developed to speed up watchlist
matching.

E already implements feature vector
indexing~\cite{DBLP:conf/birthday/Schulz13} used also for the purpose of
watchlist matching.
The watchlist clauses are inserted into an indexing data structure and 
various properties of clauses are used to prune possible subsumption candidates.
In this way, the number of possibly expensive subsumption calls is reduced.
We build upon this, and further limit the number of required subsumption checks
by using multiple indices instead of a single index.\footnote{Even with multiple watchlists, all the watchlist clauses are inserted
into a single index, and only the name of the original watchlist is additionally
stored.}

We take advantage of the fact that a clause $C$ cannot subsume a clause $D$ if
the top-level predicate symbols do not match.
In particular, $C \sqsubseteq D$ can only hold if all the predicate symbols
from $C$ also appear in $D$, because substitution can neither introduce nor
remove predicate symbols from a clause.

We define the \emph{code} of a clause $C$, denoted $\textrm{code}(C)$, as the
set of predicate symbols with their logical signs (either $+$ for positive
predicates, or $-$ for negated ones).
For example, the code of the clause ``$P(a)\lor\lnot P(b)\lor P(f(x))$'' is the set 
$\{+P,-P\}$.
The following holds because codes are preserved under
substitution.

\begin{lemma}
Given clauses $C$ and $D$, 
$C\sqsubseteq D$
implies 
$\textup{code}(C)\subseteq \textup{code}(D)$.
\end{lemma}

We create a separate index for every different clause code.
Each watchlist clause $D$ is inserted only to the index corresponding to
$\textrm{code}(D)$.
In order to check whether some clause $C$ matches a watchlist, we only need to
search in the indices whose codes are supersets of (or equal to) $\textrm{code}(C)$.
Each index is implemented using E's native feature vector indexing structure.
Evaluation of this simple indexing method is provided in Section~\ref{MultiTests}.

\section{Experiments}
\label{Eval}

This section describes the experimental evaluation
\footnote{Experiments code and data are available at  
\url{https://github.com/ai4reason/eprover-data/tree/master/TABLEAUX-19}
\newline
All experiments are run on the same hardware: Intel(R) Xeon(R) Gold 6140 CPU @ 2.30GHz with 188GB RAM.} 
of
\begin{enumerate}
    \item the improved watchlist mechanism from Section~\ref{Indexing}
    \item the watchlist selection for ENIGMAWatch from Section~\ref{ENIGMAWatch}
\end{enumerate}

\subsection{Multi-indices Subsumption Indexing Evaluation}
\label{MultiTests}

\begin{table}[t]
\setlength\tabcolsep{4pt}
\centering
\begin{minipage}{.3\linewidth}
	\includegraphics[width=3.5cm]{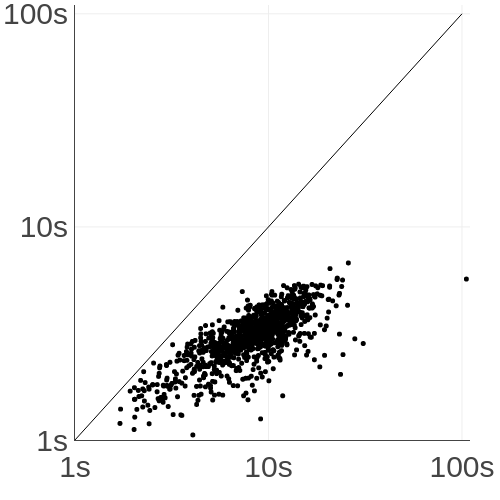}
\end{minipage}
\begin{minipage}{.38\linewidth}
   \begin{tabular}{c|ccc}
   & \multicolumn{3}{c}{runtime (left graph $\leftarrow$)} \\
   & single & multi & speedup  \\\hline
   & & &  \\[-3mm]
   avg    & 9.23s  & 3.16s & $2.9\times$   \\
   best   & 105.3s & 5.7s & $18.5\times$   \\
   worst  & 2.26s & 2.09s & $1.08\times$ 
   \end{tabular}
   \begin{tabular}{c|ccc}
   & \multicolumn{3}{c}{subsumptions (right $\rightarrow$)} \\
   & single & multi & reduction  \\\hline
   & & & \\[-3mm]
   avg   & 2328k & 52k & $44.1\times$  \\
   best  & 3059 & 1 & $3059\times$     \\
   worst & 709k & 367k & $1.9\times$
   \end{tabular}
\end{minipage}
\begin{minipage}{.3\linewidth}
	\includegraphics[width=3.5cm]{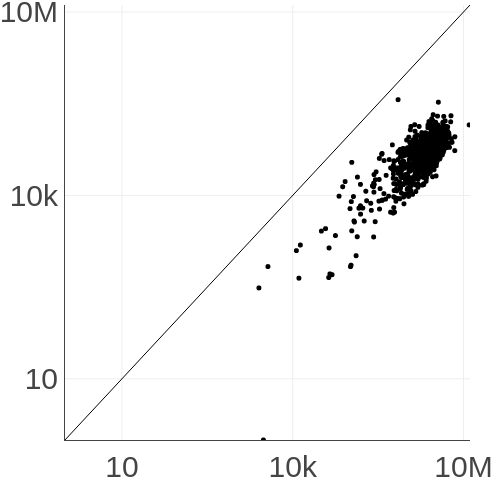}
\end{minipage}

\vspace{1mm}
\caption{\label{MultiEval}Evaluation of multi-indices subsumption indexing.}
\vspace{-8mm}
\end{table}

We propose a simple experiment to evaluate our implementation of multi-indices
subsumption indexing from Section~\ref{Indexing}.
We take a random sample of 1000 problems from the Mizar40~\cite{KaliszykU13b} data set and create a watchlist with around 60k
clauses coming from proofs of problems similar to the sample problems.
We then run E on the sample problems with a fixed limit of 1000 generated
clauses.
This gives us a measure of how fast the single-index and multi-indices
versions are, that is, how fast they can generate the first 1000 clauses.
As the watchlist indexing does not influence the proof search, both versions
process the same clauses and output the same result.
Each generated clause has to be checked for watchlist subsumption and hence the
limit on generated clauses is also the limit on different watchlist checks.
We expect the number of clause-to-clause subsumption checks to decrease with
multi-indices, as the method prunes possible subsumption candidates.

The results of the experiments are presented in Table~\ref{MultiEval}.
For each problem, we measure the runtime (left graph) and the number of different clause
subsumption calls (right graph).
The suffix ``s'' stands for seconds, ``k'' stands for thousands, and ``M'' stands for millions. 
Although subsumption is also used for purposes other than watchlist
matching, we should be able to observe a decrease in the number of calls.
Each point in the graphs corresponds to one sample problem, and is drawn at the
position $(x,y)$ corresponding to the results of single-index ($x$) and
multi-indices ($y$) versions.
Hence points below the diagonal signify an improvement.
Also note logarithmic axes.
The table shows the average improvement, and also the best
and the worst cases.
From the results, we can see that an average speed-up is almost 3 times.
Furthermore, the average reduction of subsumption calls is more than 44 times
and the number is reduced even in the worst case.

The number of watchlist clauses in the experiments was 61501, and the
multi-indices version used 11442 different indices.
This means that there were less than 6 clauses per index in average, although
the count of clauses in different indices varied from 1 to 3837.
The most crowded index was for the code $\{+=\}$, that is, for positive 
equality clauses.
Finally, 6955 indices contained only a single clause.

\subsection{Experimental Evaluation of ENIGMAWatch}
\label{ENIGMAWatchExp}

The experiments are done on a random subset of $5000$ Mizar40~\cite{KaliszykU13b} problems.
The time limit of $60$ seconds and $30000$ generated clauses is used to allow a comparison to be done without regard for the differences
in clause processing speed.
The $30000$ is approximately the average number of clauses that the baseline strategy generates
in $10$ seconds.
Table~\ref{res1} provides the evaluation of different watchlist selection mechanims
using ProofWatch (without ENIGMA) and making use of the improved watchlist
indexing. 
The last two columns show the number of problems solved by  
(1) the Baseline together with Mean, and by (2) all the five methods.
This shows the relative complementarity of the methods.
We can see that the Mean method yields the best results, reaching more than 15\%
improvement over the baseline strategy.
The Rand method is however quite competitive.

\begin{table}[t]
  \setlength\tabcolsep{4pt}
      \centering
      \begin{tabular}{c|cccc|cc}
          Baseline & Mean & Var  & Corr & Rand & Baseline $\cup$ Mean & Total\\ \hline
          1140     & 1357 & 1345 & 1337 & 1352 & 1416                 & 1483 \\
      \end{tabular}
  \caption{\label{res1}ProofWatch evaluation: Problems solved by different
  versions.}
\vspace{-4mm}
\end{table}

\begin{table}[t]
  \setlength\tabcolsep{4pt}
      \centering
      \begin{tabular}{c|ccccc|cc}
          loop & ENIGMA & Mean & Var  & Corr & Rand & ENIGMA $\cup$ Mean & Total\\ \hline
          0    & 1557   & 1694 & 1674 & 1665 & 1690 & 1830               & 1974 \\
          1    & 1776   & 1815 & 1812 & 1812 & 1847 & 1983               & 2131 \\
          2    & 1871   & 1902 & 1912 & 1882 & 1915 & 2058               & 2200 \\
          3    & 1931   & 1954 & 1946 & 1920 & 1926 & 2110               & 2227 \\
      \end{tabular}
  \caption{\label{res2}ENIGMAWatch evaluation: Problems solved and the effect of
  looping.}
\vspace{-8mm}
\end{table}

Table~\ref{res2} provides the evaluation of ENIGMAWatch and its comparison to ENIGMA. 
The experiments are done in multiple loops, where in each loop all the
proof-runs in prior loops can be used as training data. This way ENIGMA can
learn increasingly effective models.

\begin{figure}[!h]
	\begin{centering}
\vspace{-7mm}
	\includegraphics[scale=0.45]{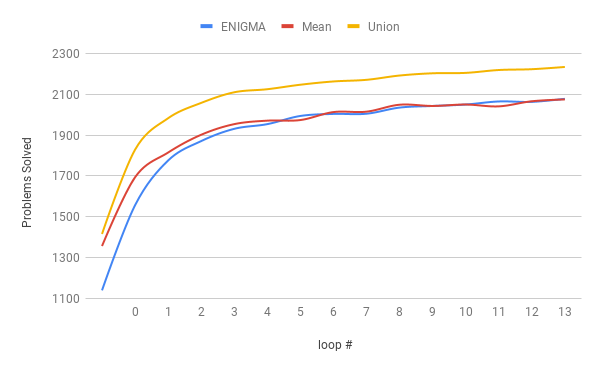}
\vspace{-7mm}
   \caption{   \label{loops}Convergence: The improvement of ENIGMA and Mean decreases over 13 loops, and their performance converges. 
        The Union is consistently ca. 150 problems better.}
	\end{centering}
\end{figure}

We can see that ENIGMAWatch can attain superior performance to ENIGMA. The relation of looping and results is interesting. The largest absolute improvement over ENIGMA is in loop 0 -- $8.8$\% by the Mean method. This however drops to $1.2$\% in loop 4.
In loops 1 and 2, Rand is the strongest, but Mean ends up
being the best in loop 3. 
In total, all the ENIGMA and ENIGMAWatch methods solve together nearly twice as many problems as the baseline strategy.
Figure \ref{loops} shows the results of running ENIGMA and Mean for 13 loops. The rate of improvement
slows down,
both methods eventually converge to a similar level of performance, and the union of the two is ca. 150 problems better. 

\vspace{-9mm}

\subsection{Training, Model Statistics and Analysis}
\label{Stats}

The XGBoost models used in our experiments are trained with a 
maximum tree depth of $9$ and $200$ rounds (which means 200
trees are learned). There are $300000$ features in the $5000$ 
problem dataset hashed into $2^{15}$ buckets. Combining
clause and conjecture features with the watchlist completion 
ratios, XGBoost makes its predictions based on $66048$ features ($2\cdot2^{15}$ plus
the count of completion ratios).

Table~\ref{xgbstats} provides various training and model statistics of the ENIGMA
and ENIGMAWatch models and their loops.
The columns ``Pos. Acc.'' and ``Neg. Acc.'' describe the training accuracy of
the models on positive and negative training examples.
The column ``Features'' presents the number of features referenced in the decision
trees.
We see that the models  use a small fraction of all the $66048$ available features.
The column ``Watchlist F.'' provides the number of watchlist features out of all
the used features.
Finally, ``Train Size'' and ``Train Time'' specify the size of the input
training file (in GB) and training times (in minutes).
The XGBoost models after the training are smaller than $4$ MB.

\begin{table}[t]
\begin{small}
  \setlength\tabcolsep{4pt}
      \centering
      \begin{tabular}{c|rrrrrr}
          Model & Pos. Acc. & Neg. Acc. & Features & Watchlist F. & Train Size & Train Time \\ \hline 
          ENIGMA0 & 99.12\% & 92.16\% & 5061 & 0 & 0.4 GB   &  14min\\
          ENIGMA1 & 97.39\% & 86.82\% & 7071 & 0 & 0.8 GB   &  31min \\
          ENIGMA2 & 96.13\% & 83.92\% & 8089 & 0 & 1.4 GB   &  55min\\
          ENIGMA3 & 95.39\% & 82.5 \% & 8662 & 0 & 2.0 GB   &  85min \\\hline
          Mean0   & 99.05\% & 92.59\% & 5424 & 308 & 2.9 GB &  19min\\
          Mean1   & 96.92\% & 88.16\% & 6950 & 316 & 6.2 GB &  29min\\
          Mean2   & 95.75\% & 86.46\% & 7809 & 331 & 9.6 GB &  38min\\
          Mean3   & 95.04\% & 85.24\% & 8313 & 330 & 13.0 GB &  39min\\
      \end{tabular}
    \caption{\label{xgbstats}ENIGMA and ENIGMAWatch: Model and training statistics.}
\vspace{-6mm}
\end{small}
\end{table}

\begin{figure}[!h]
	\begin{centering}
\vspace{-8mm}
	\includegraphics[width=0.9\textwidth]{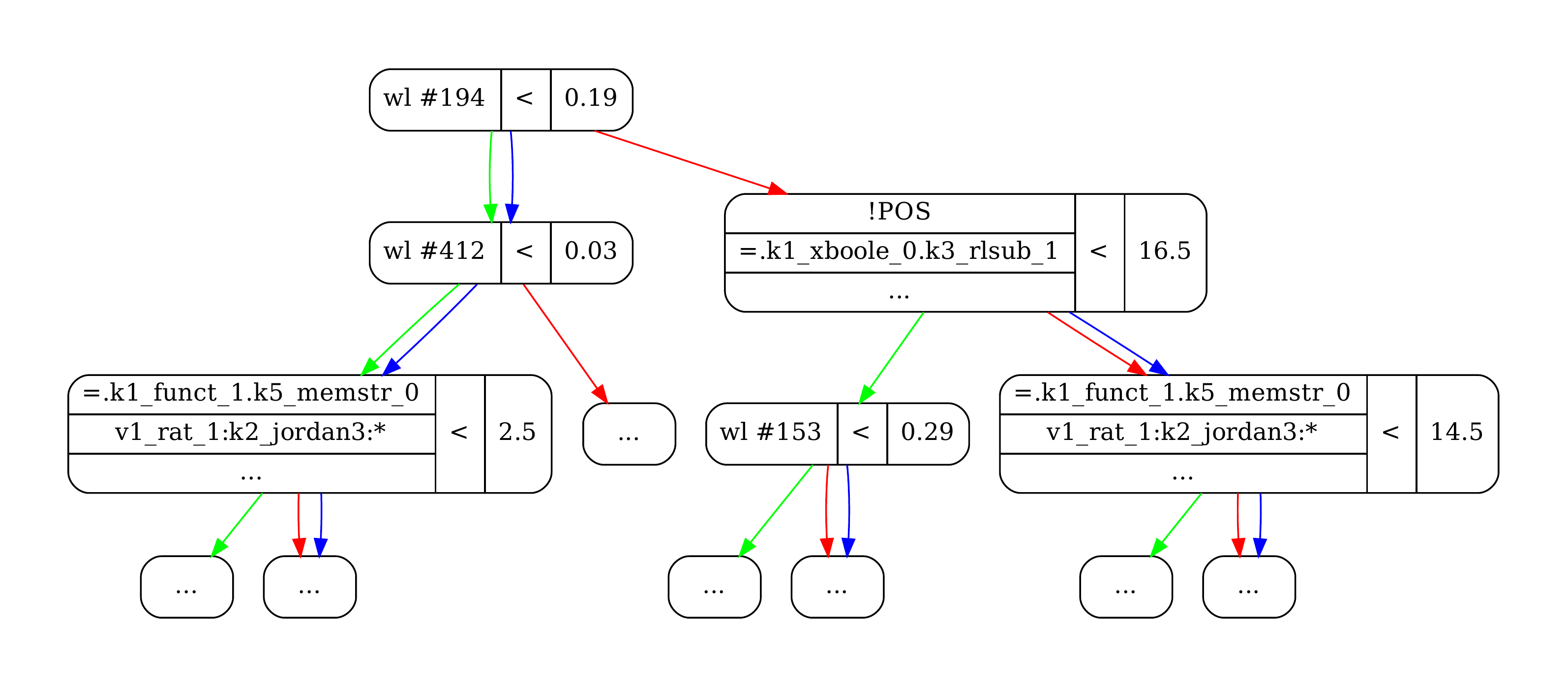}
\vspace{-8mm}
   \caption{   \label{Treefig}Example of an XGBoost decision tree.}
	\end{centering}
\end{figure}

We can see that the accuracy decreases with the increase of the training data size, but the number of
theorems proved increases. About $62\%$ of the watchlists are judged
as useful by XGBoost and used in the decision trees. 
Figure~\ref{Treefig} shows the root of the first decision tree of the Mean model in
loop 3. 
Green means "yes" (the condition holds), red means "no", and blue means that the
feature is not present.
The multi line box is a (shortened) bucket of features, and single line boxes correspond to
watchlists ($\#194$, etc.).  
We can see that ENIGMAWatch uses a watchlist feature for the very first decision when judging newly generated clauses.
This shows that the features that characterize the evolving proof state are indeed considered very significant
by the methods that automatically learn given clause  guidance.

\section{Conclusion and Future Work}
We have produced and evaluated the first practically usable version
of the ENIGMAWatch system which can now be efficiently used over large
mathematical datasets. The previous experiments with the first
prototype on the small MPTP
Challenge~\cite{LPAR-IWIL2018:ProofWatch_Meets_ENIGMA_First}
demonstrated that ENIGMAWatch can find proofs faster
(in terms of how many processed
clauses are needed). 
The work presented here shows that with improved subsumption indexing, feature hashing, 
and suitable global watchlist selection, 
ENIGMAWatch outperforms ENIGMA
on the large Mizar40 dataset. In particular, ENIGMAWatch significantly outperforms both ProofWatch and ENIGMA
when used without looping. With several MaLARea-style~\cite{Urban07,US+08-long} iterations of proving and learning, the difference to ENIGMA gets smaller, however the two methods are still
quite complementary, providing solutions to a large number of different problems. In total, all the ENIGMA and ENIGMAWatch methods (Table~\ref{res2}) together solve almost twice as many problems as the baseline strategy after four iterations of learning and proving.

The system is ready to be used on hard problems and to expand the set of
Mizar problems for which an ATP proof has been found. Future work
includes refining the watchlist selection, defining more sophisticated
methods of computing the proof completion ratios, analyzing the
learned decision tree models to see which watchlists are the most
useful, and also defining further and more abstract meaningful
representations and embeddings of saturation-style proof search.

\bibliographystyle{abbrv}
\bibliography{ate11.bib,stsbib.bib}

\begin{thebibliography}{10}

\bibitem{abs-1108-3446}
J.~Alama, T.~Heskes, D.~K\"{u}hlwein, E.~Tsivtsivadze, and J.~Urban.
\newblock Premise selection for mathematics by corpus analysis and kernel
  methods.
\newblock {\em J. Autom. Reasoning}, 52(2):191--213, 2014.

\bibitem{IrvingSAECU16}
A.~A. Alemi, F.~Chollet, N.~E{\'{e}}n, G.~Irving, C.~Szegedy, and J.~Urban.
\newblock {DeepMath} - deep sequence models for premise selection.
\newblock In D.~D. Lee, M.~Sugiyama, U.~V. Luxburg, I.~Guyon, and R.~Garnett,
  editors, {\em Advances in Neural Information Processing Systems 29: Annual
  Conference on Neural Information Processing Systems 2016, December 5-10,
  2016, Barcelona, Spain}, pages 2235--2243, 2016.

\bibitem{BlanchetteGKKU16}
J.~C. Blanchette, D.~Greenaway, C.~Kaliszyk, D.~K{\"{u}}hlwein, and J.~Urban.
\newblock A learning-based fact selector for {Isabelle/HOL}.
\newblock {\em J. Autom. Reasoning}, 57(3):219--244, 2016.

\bibitem{DBLP:journals/jar/BridgeHP14}
J.~P. Bridge, S.~B. Holden, and L.~C. Paulson.
\newblock Machine learning for first-order theorem proving - learning to select
  a good heuristic.
\newblock {\em J. Autom. Reasoning}, 53(2):141--172, 2014.

\bibitem{DBLP:conf/kdd/ChenG16}
T.~Chen and C.~Guestrin.
\newblock Xgboost: {A} scalable tree boosting system.
\newblock In {\em {KDD}}, pages 785--794. {ACM}, 2016.

\bibitem{DBLP:journals/corr/abs-1903-03182}
K.~Chvalovsk{\'{y}}, J.~Jakubuv, M.~Suda, and J.~Urban.
\newblock {ENIGMA-NG:} efficient neural and gradient-boosted inference guidance
  for {E}.
\newblock {\em CoRR}, abs/1903.03182, 2019.

\bibitem{DenzingerFGS99}
J.~Denzinger, M.~Fuchs, C.~Goller, and S.~Schulz.
\newblock {Learning from Previous Proof Experience}.
\newblock Technical Report AR99-4, Institut f{\"u}r Informatik, Technische
  Universit{\"a}t M{\"u}nchen, 1999.

\bibitem{DBLP:conf/ogai/ErtelSS89}
W.~Ertel, J.~Schumann, and C.~B. Suttner.
\newblock Learning heuristics for a theorem prover using back propagation.
\newblock In J.~Retti and K.~Leidlmair, editors, {\em 5. {\"{O}}sterreichische
  Artificial Intelligence-Tagung, Igls, Tirol, 28. bis 30. September 1989,
  Proceedings}, volume 208 of {\em Informatik-Fachberichte}, pages 87--95.
  Springer, 1989.

\bibitem{DBLP:conf/cade/FarberB16}
M.~F{\"{a}}rber and C.~E. Brown.
\newblock Internal guidance for satallax.
\newblock In N.~Olivetti and A.~Tiwari, editors, {\em Automated Reasoning - 8th
  International Joint Conference, {IJCAR} 2016, Coimbra, Portugal, June 27 -
  July 2, 2016, Proceedings}, volume 9706 of {\em Lecture Notes in Computer
  Science}, pages 349--361. Springer, 2016.

\bibitem{hh4h4}
T.~Gauthier and C.~Kaliszyk.
\newblock Premise selection and external provers for {HOL4}.
\newblock In {\em Certified Programs and Proofs (CPP'15)}, LNCS. Springer,
  2015.
\newblock \url{http://dx.doi.org/10.1145/2676724.2693173}.

\bibitem{DBLP:conf/itp/GoertzelJ0U18}
Z.~Goertzel, J.~Jakubuv, S.~Schulz, and J.~Urban.
\newblock {ProofWatch}: Watchlist guidance for large theories in {E}.
\newblock In J.~Avigad and A.~Mahboubi, editors, {\em Interactive Theorem
  Proving - 9th International Conference, {ITP} 2018, Held as Part of the
  Federated Logic Conference, FloC 2018, Oxford, UK, July 9-12, 2018,
  Proceedings}, volume 10895 of {\em Lecture Notes in Computer Science}, pages
  270--288. Springer, 2018.

\bibitem{LPAR-IWIL2018:ProofWatch_Meets_ENIGMA_First}
Z.~Goertzel, J.~Jakubuv, and J.~Urban.
\newblock {ProofWatch} meets {ENIGMA}: First experiments.
\newblock In G.~Barthe, K.~Korovin, S.~Schulz, M.~Suda, G.~Sutcliffe, and
  M.~Veanes, editors, {\em LPAR-22 Workshop and Short Paper Proceedings},
  volume~9 of {\em Kalpa Publications in Computing}, pages 15--22. EasyChair,
  2018.

\bibitem{DBLP:conf/gcai/2015}
G.~Gottlob, G.~Sutcliffe, and A.~Voronkov, editors.
\newblock {\em Global Conference on Artificial Intelligence, {GCAI} 2015,
  Tbilisi, Georgia, October 16-19, 2015}, volume~36 of {\em EPiC Series in
  Computing}. EasyChair, 2015.

\bibitem{mizar-in-a-nutshell}
A.~Grabowski, A.~Korni{\l}owicz, and A.~Naumowicz.
\newblock {M}izar in a nutshell.
\newblock {\em J. Formalized Reasoning}, 3(2):153--245, 2010.

\bibitem{JakubuvU18a}
J.~Jakub\r{u}v and J.~Urban.
\newblock Hierarchical invention of theorem proving strategies.
\newblock {\em {AI} Commun.}, 31(3):237--250, 2018.

\bibitem{JakubuvU17a}
J.~Jakubuv and J.~Urban.
\newblock {ENIGMA:} efficient learning-based inference guiding machine.
\newblock In H.~Geuvers, M.~England, O.~Hasan, F.~Rabe, and O.~Teschke,
  editors, {\em Intelligent Computer Mathematics - 10th International
  Conference, {CICM} 2017, Edinburgh, UK, July 17-21, 2017, Proceedings},
  volume 10383 of {\em Lecture Notes in Computer Science}, pages 292--302.
  Springer, 2017.

\bibitem{JakubuvU18}
J.~Jakubuv and J.~Urban.
\newblock Enhancing {ENIGMA} given clause guidance.
\newblock In F.~Rabe, W.~M. Farmer, G.~O. Passmore, and A.~Youssef, editors,
  {\em Intelligent Computer Mathematics - 11th International Conference, {CICM}
  2018, Hagenberg, Austria, August 13-17, 2018, Proceedings}, volume 11006 of
  {\em Lecture Notes in Computer Science}, pages 118--124. Springer, 2018.

\bibitem{DBLP:journals/corr/abs-1904-01677}
J.~Jakubuv and J.~Urban.
\newblock Hammering {Mizar} by learning clause guidance.
\newblock {\em CoRR}, abs/1904.01677, 2019.

\bibitem{holyhammer}
C.~Kaliszyk and J.~Urban.
\newblock Learning-assisted automated reasoning with {F}lyspeck.
\newblock {\em J. Autom. Reasoning}, 53(2):173--213, 2014.

\bibitem{KaliszykU15}
C.~Kaliszyk and J.~Urban.
\newblock {FEMaLeCoP}: Fairly efficient machine learning connection prover.
\newblock In M.~Davis, A.~Fehnker, A.~McIver, and A.~Voronkov, editors, {\em
  Logic for Programming, Artificial Intelligence, and Reasoning - 20th
  International Conference, {LPAR-20} 2015, Suva, Fiji, November 24-28, 2015,
  Proceedings}, volume 9450 of {\em Lecture Notes in Computer Science}, pages
  88--96. Springer, 2015.

\bibitem{KaliszykU13b}
C.~Kaliszyk and J.~Urban.
\newblock {MizAR 40 for Mizar 40}.
\newblock {\em J. Autom. Reasoning}, 55(3):245--256, 2015.

\bibitem{KaliszykUMO18}
C.~Kaliszyk, J.~Urban, H.~Michalewski, and M.~Ols{\'{a}}k.
\newblock Reinforcement learning of theorem proving.
\newblock In {\em Advances in Neural Information Processing Systems 31: Annual
  Conference on Neural Information Processing Systems 2018, NeurIPS 2018, 3-8
  December 2018, Montr{\'{e}}al, Canada.}, pages 8836--8847, 2018.

\bibitem{DBLP:conf/ijcai/KaliszykUV15}
C.~Kaliszyk, J.~Urban, and J.~Vyskocil.
\newblock Efficient semantic features for automated reasoning over large
  theories.
\newblock In {\em {IJCAI}}, pages 3084--3090. {AAAI} Press, 2015.

\bibitem{KinyonVV13}
M.~K. Kinyon, R.~Veroff, and P.~Vojtechovsk{\'{y}}.
\newblock Loops with abelian inner mapping groups: An application of automated
  deduction.
\newblock In M.~P. Bonacina and M.~E. Stickel, editors, {\em Automated
  Reasoning and Mathematics - Essays in Memory of William W. McCune}, volume
  7788 of {\em LNCS}, pages 151--164. Springer, 2013.

\bibitem{Vampire}
L.~Kov{\'a}cs and A.~Voronkov.
\newblock First-order theorem proving and {V}ampire.
\newblock In N.~Sharygina and H.~Veith, editors, {\em CAV}, volume 8044 of {\em
  LNCS}, pages 1--35. Springer, 2013.

\bibitem{LoosISK17}
S.~M. Loos, G.~Irving, C.~Szegedy, and C.~Kaliszyk.
\newblock Deep network guided proof search.
\newblock In T.~Eiter and D.~Sands, editors, {\em LPAR-21, 21st International
  Conference on Logic for Programming, Artificial Intelligence and Reasoning,
  Maun, Botswana, May 7-12, 2017}, volume~46 of {\em EPiC Series in Computing},
  pages 85--105. EasyChair, 2017.

\bibitem{MW:JAR-97}
W.~McCune and L.~Wos.
\newblock {Otter: The CADE-13 Competition Incarnations}.
\newblock {\em Journal of Automated Reasoning}, 18(2):211--220, 1997.
\newblock Special Issue on the CADE 13 ATP System Competition.

\bibitem{McCune:WWW-2008}
W.~W. McCune.
\newblock {Prover9 and Mace4}.
\newblock \url{http://www.cs.unm.edu/~mccune/prover9/}, 2005--2010.
\newblock (acccessed 2016-03-29).

\bibitem{PiotrowskiU18}
B.~Piotrowski and J.~Urban.
\newblock {ATPboost}: Learning premise selection in binary setting with {ATP}
  feedback.
\newblock In D.~Galmiche, S.~Schulz, and R.~Sebastiani, editors, {\em Automated
  Reasoning - 9th International Joint Conference, {IJCAR} 2018, Held as Part of
  the Federated Logic Conference, FloC 2018, Oxford, UK, July 14-17, 2018,
  Proceedings}, volume 10900 of {\em Lecture Notes in Computer Science}, pages
  566--574. Springer, 2018.

\bibitem{Polikar06}
R.~Polikar.
\newblock {Ensemble based systems in decision making}.
\newblock {\em Circuits and Systems Magazine, IEEE}, 6(3):21--45, 2006.

\bibitem{SchaferS15}
S.~Sch{\"{a}}fer and S.~Schulz.
\newblock Breeding theorem proving heuristics with genetic algorithms.
\newblock In Gottlob et~al. \cite{DBLP:conf/gcai/2015}, pages 263--274.

\bibitem{DBLP:books/daglib/0002958}
S.~Schulz.
\newblock {\em Learning search control knowledge for equational deduction},
  volume 230 of {\em {DISKI}}.
\newblock Infix Akademische Verlagsgesellschaft, 2000.

\bibitem{Sch02-AICOMM}
S.~Schulz.
\newblock {E - A Brainiac Theorem Prover}.
\newblock {\em AI Commun.}, 15(2-3):111--126, 2002.

\bibitem{DBLP:conf/birthday/Schulz13}
S.~Schulz.
\newblock Simple and efficient clause subsumption with feature vector indexing.
\newblock In {\em Automated Reasoning and Mathematics}, volume 7788 of {\em
  Lecture Notes in Computer Science}, pages 45--67. Springer, 2013.

\bibitem{Schulz13}
S.~Schulz.
\newblock System description: {E} 1.8.
\newblock In K.~L. McMillan, A.~Middeldorp, and A.~Voronkov, editors, {\em
  LPAR}, volume 8312 of {\em LNCS}, pages 735--743. Springer, 2013.

\bibitem{Urban06}
J.~Urban.
\newblock {MPTP} 0.2: Design, implementation, and initial experiments.
\newblock {\em J. Autom. Reasoning}, 37(1-2):21--43, 2006.

\bibitem{Urban07}
J.~Urban.
\newblock {MaLARea}: a metasystem for automated reasoning in large theories.
\newblock In G.~Sutcliffe, J.~Urban, and S.~Schulz, editors, {\em ESARLT},
  volume 257 of {\em CEUR Workshop Proceedings}. CEUR-WS.org, 2007.

\bibitem{blistr}
J.~Urban.
\newblock {BliStr: The Blind Strategymaker}.
\newblock In Gottlob et~al. \cite{DBLP:conf/gcai/2015}, pages 312--319.

\bibitem{UrbanMPTPChallenge}
J.~Urban and G.~Sutcliffe.
\newblock {ATP} cross-verification of the {Mizar MPTP Challenge} problems.
\newblock In N.~Dershowitz and A.~Voronkov, editors, {\em Logic for
  Programming, Artificial Intelligence, and Reasoning}, pages 546--560, Berlin,
  Heidelberg, 2007. Springer Berlin Heidelberg.

\bibitem{US+08-long}
J.~Urban, G.~Sutcliffe, P.~Pudl{\'a}k, and J.~Vysko\v{c}il.
\newblock {MaLARea SG1 - Machine Learner for Automated Reasoning with Semantic
  Guidance}.
\newblock In A.~Armando, P.~Baumgartner, and G.~Dowek, editors, {\em IJCAR},
  volume 5195 of {\em LNCS}, pages 441--456. Springer, 2008.

\bibitem{UrbanVS11}
J.~Urban, J.~Vysko\v{c}il, and P.~\v{S}t\v{e}p{\'a}nek.
\newblock {MaLeCoP}: Machine learning connection prover.
\newblock In K.~Br{\"u}nnler and G.~Metcalfe, editors, {\em TABLEAUX}, volume
  6793 of {\em LNCS}, pages 263--277. Springer, 2011.

\bibitem{Veroff:JAR-1996}
R.~Veroff.
\newblock Using hints to increase the effectiveness of an automated reasoning
  program: Case studies.
\newblock {\em Journal of Automated Reasoning}, 16(3):223--239, 1996.

\end{thebibliography}
\end{document}